\documentclass[10pt,twocolumn,letterpaper]{article}

\usepackage{iccv}
\usepackage{times}
\usepackage{epsfig}
\usepackage{graphicx}
\usepackage{amsmath}
\usepackage{amssymb}

\usepackage[accsupp]{axessibility}  
\newcommand{\smallsec}[1]{\vspace{.14cm} \noindent {\bf #1.}}

\newcommand{\gendered}{\textbf{Gendered}}
\newcommand{\control}{{\textbf{Control}}}
\newcommand{\scratch}{{\textbf{Scratch}}}
\usepackage{tabularx}
\usepackage{multirow}

\usepackage[breaklinks=true,bookmarks=false]{hyperref}

\iccvfinalcopy 

\def\iccvPaperID{3956} 

\ificcvfinal\fi

\begin{document}

\title{Overwriting Pretrained Bias with Finetuning Data}

\author{Angelina Wang\\
Princeton University\\
{\tt\small angelina.wang@princeton.edu}
\and
Olga Russakovsky\\
Princeton University\\
{\tt\small olgarus@princeton.edu}
}

\maketitle
\ificcvfinal\thispagestyle{empty}\fi

\begin{abstract}
Transfer learning is beneficial by allowing the expressive features of models pretrained on large-scale datasets to be finetuned for the target task of smaller, more domain-specific datasets. However, there is a concern that these pretrained models may come with their own biases which would propagate into the finetuned model. In this work, we investigate bias when conceptualized as both \textit{spurious correlations} between the target task and a sensitive attribute as well as \textit{underrepresentation} of a particular group in the dataset. Under both notions of bias, we find that (1) models finetuned on top of pretrained models can indeed inherit their biases, but (2) this bias can be corrected for through relatively minor interventions to the finetuning dataset, and often with a negligible impact to performance. Our findings imply that careful curation of the finetuning dataset is important for reducing biases on a downstream task, and doing so can even compensate for bias in the pretrained model.
\end{abstract}
\section{Introduction}

The current paradigm in machine learning typically involves using an off-the-shelf pretrained model that has been trained on a large-scale dataset, and then finetuning it on a smaller, application-specific dataset. This transfer learning is especially common for high-dimensional data like images and language~\cite{girshick2014rich, donahue2014generic, zeiler2014understanding}. However, large-scale datasets have been criticized for their biases~\cite{prabhu21pyrrhic, revisetool_extended, birhane2021multimodal, crawford2019excavating}, which leaves open the concern that models pretrained on such datasets may carry biases over into the finetuned model. On the other hand, pretraining has been shown to confer benefits in model robustness and uncertainty estimation~\cite{hendrycks2019robustness}, so there is also the potential that pretrained models can reduce downstream biases by being more regularized or resistant to spurious correlations~\cite{bommasani2022homogenization}. In our work, we investigate the implications of bias in pretrained models for the downstream finetuning task, and provide actionable insights on how to counteract this.

We have been deliberately vague thus far on what we mean by ``bias.'' In this work, we operationalize bias in two ways based on what has been found thus far to be problematic in image features: \textit{spurious correlations} between a sensitive attribute and target task~\cite{zhao2017menshop, wang2021biasamp, singh2020context, wang2019balanced} and reduced performance from \textit{underrepresentation}~\cite{buolamwini2018gendershades,devries19everyone, shankar2017geodiversity}. 

The topic of whether pretrained biases matter in finetuning is often assumed to be obvious, with contradictory arguments containing intuitively plausible explanations on both sides of the debate: that it does matter because the pretrained model brings biased features~\cite{salman2022transfer}, or it does not because finetuning data will overwrite any pretrained biases~\cite{cao2022intrinsic, goldfarb-tarrant2021intrinsic, steed2022transfer}. Due to this uncertainty, it is not clear how to react to biases found in the features of pretrained models~\cite{steed2021representations, sirotkin2022selfsupervised, goyal2022fairfeatures}.
Of course, there is not a singular binary answer, as much is dependent upon the particulars of the training task. However, we bring much-needed clarity to the space for computer vision tasks, and give advice about bias transference from using pretrained models. 

\begin{figure}[t!]
    \centering
    \includegraphics[width=0.47\textwidth]{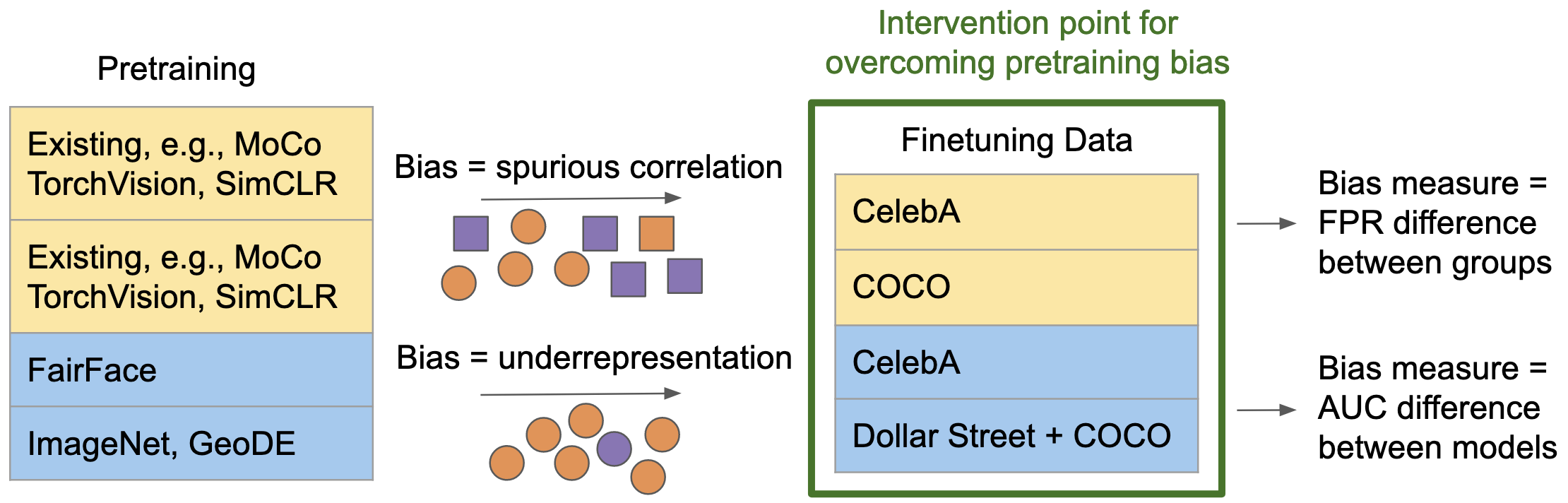}
    \caption{We explore bias transference from pretrained to finetuned models in two forms in this work: spurious correlations and underrepresentation. We find that intervening on the finetuning data allows us to overcome bias from pretraining, often without compromising on performance.
    }
    \label{fig:fig1}
\end{figure}

In this work we study the two notions of bias, \textit{spurious correlations} and \textit{underrepresentation}, by finetuning a variety of different pretrained models (Fig.~\ref{fig:fig1}). For each notion of bias, we first show our results on the CelebA dataset~\cite{liu2015celeba}. Then for spurious correlations we investigate the more complex COCO dataset~\cite{li2014coco} using real-world popular pretrained models (e.g., \textbf{MoCo}~\cite{he2020moco}, \textbf{SimCLR}~\cite{chen2020simclr}). For underrepresentation, we look to the Dollar Street dataset~\cite{rojas2022dollarstreet} using pretrained models of our own design to test for specific hypotheses.
On both forms of bias we find the following: 1) models finetuned on top of pretrained models can inherit their biases (for spurious correlations, this is especially true if the correlation level is high, the salience of the bias signal is high relative to the true task signal, and/or the number of finetuning samples is low); 2) this bias can be relatively easily corrected for by curating the distribution of the finetuning dataset, \textit{with a negligible impact to performance}.
For example on CelebA, we find that by manipulating the strength of the spurious correlation in the finetuning dataset from 20\% to 30\%, we can retain the same high performance from using a biased pretrained model, but cut the amount of bias almost in half.
The implications for this are significant: practitioners can use the pretrained model that lends the best performance in most cases so long as they appropriately curate the finetuning dataset, and thus get the best of both worlds in terms of performance and fairness. This means that significant consideration and effort needs to be spent on the curation of finetuning datasets, in a way that may not necessarily reflect the distribution of the test set, in order to ``correct'' for the biases of the pretrained model.\footnote{Code at \url{https://github.com/princetonvisualai/overcoming-pretraining-bias}.}




\section{Related Work}
Rather than exploring all viable algorithmic bias mitigation strategies, we focus in this work on studying the already sufficiently-complex effect of the finetuning dataset's composition. As datasets grow increasingly larger in size and harder to control, we position our work as directing attention to the finetuning dataset (as opposed to the pretraining dataset) as the better site for investigation and mitigation.

\smallsec{Benefits of Pretraining}
Pretrained features can be learned through methods like unsupervised~\cite{chen2020simclr, he2020moco}, self-supervised~\cite{newell2020selfsupervised, el-nouby2021selfsupervised}, or supervised learning~\cite{pytorch}.
While there are indisputable time- and compute-saving benefits to transfer learning, He et al.~\cite{he2019pretraining} showed that sometimes models trained from scratch can match the performance of models finetuned from pretrained weights. However, even in those cases, Hendrycks et al.~\cite{hendrycks2019robustness} finds that the finetuned model has superior robustness and uncertainty estimates. 
In our work, we try to understand whether pretrained models confer any other such benefits or harms in terms of bias. 


\smallsec{Identifying Pretraining Bias}
Prior works have sought to measure the fairness of image features~\cite{goyal2022fairfeatures, steed2021representations, sirotkin2022selfsupervised}, the concern being that any biases in image features may propagate into predictions based on such features. Goyal et al.~\cite{goyal2022fairpretrain} report that pretrained features trained without supervision are more robust and fair than those with supervision, and Sirotkin et al.~\cite{sirotkin2022selfsupervised} find that within self-supervised models, contrastive losses lead to more biases---however, neither of these works measure bias on a downstream task.\footnote{While Goyal et al.~\cite{goyal2022fairpretrain} finetune their pretrained models on ImageNet for certain measurements, this is only to be able to extract labels from a self-supervised model and ultimately assess those features.} So, while the pretrained features may be biased, it still does not tell us the implications for a downstream task.

Prior work in NLP has measured the correlation of ``intrinsic'' metrics like biases in the word embedding to ``extrinsic'' metrics like biases on the downstream task which uses these word embeddings, finding little correlation between these metrics~\cite{goldfarb-tarrant2021intrinsic, cao2022intrinsic}. Other work in NLP has found that the distribution of the finetuning dataset matters more than the pretraining dataset in terms of bias~\cite{steed2022transfer, solaiman2021palms}. We find similar results in our study on pretrained image models, and give advice on how to correct for this. There is also work beginning to study the multimodal pretraining biases in the vision language domain~\cite{srinivasan2022worst}.

\smallsec{Bias Transferrence in Vision}
Salman et al. \cite{salman2022transfer} similarly study bias transference, but focus on an operationalization of bias akin to backdoor attacks~\cite{gu2017badnets}. In their work and ours, we find that biases in pretrained models can propagate to finetuned models, but in our work we take this a step further by productively demonstrating that relatively minor interventions on the finetuning dataset can counteract this. While they find that in certain settings the biases in pretrained models persist even when the downstream dataset does not contain such biases, this different conclusion is likely due to their freezing of model layers.
Kirichenko et al. \cite{kirichenko2023spurious} find that re-training just the last layer of a network can help models overcome spurious correlations. This matches the setting of one of our two conceptualizations of bias, and we add new findings when bias is underrepresentation. While their claim is stronger because they only retrain the last layer whereas we retrain the entire network, the setting we consider is very different because in our scenarios the available number of finetuning samples is relatively small, e.g., $128$ or $1024$ on CelebA, whereas in their comparable experiments they have access to the entire CelebA dataset ($130,216$ samples). We find our setting of smaller finetuning datasets to be more realistic to many real-world scenarios.

\section{Preliminaries}
In this work, we will be conceptualizing ``bias'' in its two most studied forms in computer vision: \textit{spurious correlations} and \textit{underrepresentation}. Each of our experimental setups will consist first of a \textit{pretraining} stage on the upstream task and then a \textit{finetuning} stage on the downstream task. The \textit{pretraining} stage can include simply instantiating a randomly initialized model (\scratch{}), instantiating the model weights to be that of an existing pretrained model (e.g., \textbf{SimCLR}~\cite{chen2020simclr}, \textbf{MoCo}~\cite{he2020moco}), or pretraining our own model (e.g., \gendered{}). All pretrained models will be denoted in bold text.
In the \textit{finetuning} stage we train all layers of our model on the downstream task. 
We do not experiment with any freezing of layers, and thus our results serve as the upper bound for the effect that finetuning can have.

\subsection{Datasets}
\label{sec:prelim_datasets}
We use six datasets in this work. They serve as training data for our upstream pretrained model, training data for our downstream finetuned model, or both.
We use gender as our sensitive attribute for our analysis on spurious correlations, and use the terms \texttt{Men} and \texttt{Women} when referring to the annotated gender groups.\footnote{Much fairness work is limited by the availability of sensitive attribute annotations~\cite{andrus2021measure}, and our annotations generally treat gender as binary, a schema which harmfully alienates and erases different communities~\cite{scheuerman2021gender}.} Our datasets are:
\begin{itemize}
    \item CelebA~\cite{liu2015celeba}: 39 attributes such as \texttt{Brown Hair} and \texttt{High Cheekbones}; we use the label of \texttt{Male} for our \texttt{Men} group and not-\texttt{Male} for our \texttt{Women} group.
    \item FairFace~\cite{karkkainen2021fairface}: racial attributes such as \texttt{White}; we use the labels of \texttt{Male} for \texttt{Men} and \texttt{Female} for \texttt{Women}.
    \item COCO~\cite{li2014coco}: 80 objects such as \texttt{Toothbrush} and \texttt{Oven}; we follow from prior work~\cite{zhao2017menshop, zhao2021caption} and derive gender labels based on the presence of gendered words in the captions. Even though COCO is most often a multi-label task, in some experiments we perform binary prediction on one label at a time to better isolate and manipulate the correlation of that particular object. Most images are from the Global North~\cite{devries19everyone}.
    \item ImageNet~\cite{imagenet}: 200 objects such as \texttt{Fence} and \texttt{Barn}; most images are from the Global North~\cite{devries19everyone}.
    \item DollarStreet~\cite{rojas2022dollarstreet, devries19everyone}: 135 objects such as \texttt{Books} and \texttt{Motorcycles}, of which 15 map onto COCO object labels; this dataset is more geographically diverse than, e.g., COCO and ImageNet.
    \item GeoDE~\cite{ramaswamy2023geode}: 40 objects such as \texttt{Bicycle} and \texttt{Jug}; deliberately collected to be geographically diverse.
\end{itemize}

A summary of how each dataset is used is in Fig.~\ref{fig:fig1}.

\subsection{Implementation Details}
We use a ResNet50~\cite{he2016resnet} with SGD for all of our experiments. The images are resized to be 224 by 224, normalized to ImageNet's mean pixel values, and randomly flipped horizontally for data augmentation. The particular hyperparameters chosen for finetuning are impactful~\cite{li2020hyperparameters}, so we perform grid search for the lowest validation loss across the following hyperparameters: learning rate in $\{.1, .05, .01, .005, .001\}$ and weight decay in $\{0., .0001, .0005, .001\}$.\footnote{For similar settings, e.g., same task with a different dataset distribution, we reuse the same hyperparameters in order to save computation time.}
We train five random runs of every model to generate 95\% confidence intervals. 
When we need discrete predictions, i.e., $\{0, 1\}$, as opposed to continuous ones, i.e., $[0, 1]$, we pick the classification threshold for each label to be well-calibrated such that the percentage of predicted positive labels is the same as the true percentage.


\section{Bias as Spurious Correlation}

We begin our analysis by considering bias in the form of spurious correlations between the target label and a sensitive attribute which is predictive on the training set but not necessarily so on the test set~\cite{zhao2017menshop, wang2021biasamp, singh2020context, wang2019balanced, sagawa2020spurious}. We first demonstrate that finetuned models can inherit bias of this form from pretrained models on the CelebA dataset (Sec.~\ref{sec:sc_inherit}). In other words, when the presence of a target label (e.g., \texttt{Eyebags}) may be erroneously learned to be correlated with gender, such that the classifier over-predicts \texttt{Eyebags} (i.e., a false positive) on images with \texttt{Women}. We thus measure the bias of spurious correlations as false positive rate (FPR) difference between \texttt{Women} and \texttt{Men}.\footnote{Though often called equal opportunity, we will stay away from this term to not detract from its namesake~\cite{hardt2016equalopp}, which is focused on allocational harms, when our application likely leads to representational harms~\cite{barocas2017problem}.} 
We show that pretrained bias is especially likely to affect the downstream task when the finetuning dataset has a high correlation with the sensitive attribute, low salience relative to the sensitive attribute, and/or a low number of finetuning samples.

Then we examine bias mitigation through finetuning on different distributions of the downstream dataset (Sec.~\ref{sec:sc_correct}). 
We show on two domains (CelebA attribute classification~\cite{liu2015celeba} and COCO object recognition~\cite{li2014coco}) that interventions on the finetuning dataset can mitigate much of the bias while retaining the performance gains of a biased pretrained model.


\subsection{Finetuned models inherit spurious correlations from pretrained models}
\label{sec:sc_inherit}

\smallsec{Initial Look}
To start off, we want a set of pretrained models of differing biases to finetune on a set of downstream tasks where we can measure bias (i.e., FPR difference between \texttt{Women} and \texttt{Men}) in order to assess whether it makes a difference which pretrained model was used.
For our downstream tasks, we perform binary classification on 11 different CelebA attributes.\footnote{We selected the 11 attributes as follows: taking gender as our sensitive attribute, we are left with 39 in CelebA. Following from Ramaswamy et al.~\cite{ramaswamy2021latent}, we winnow this down to the 26 that do not suffer from severe data imbalance (i.e., positive label rate between 5-95\%), then the 20 that are consistently labeled (e.g., \texttt{Black} \texttt{Hair} and not \texttt{Big} \texttt{Nose}), then the 14 that look the same between people of different genders (e.g., \texttt{Eyeglasses} and not \texttt{Young}). From here, we use three as to train our \control{} models, and are left with 11 to use as downstream tasks.}
To understand whether the bias of a pretrained model (i.e., learned correlation between gender and target attribute) will affect the bias on a finetuning task (i.e., applied correlation of gender and target attribute during inference), we compare a pretrained model of our own design with high bias, called \gendered{}, to three of lesser bias, called \control{}.
We then finetune all pretrained models and compare their downstream FPR gender difference. 
We create \gendered{} by training a model from scratch to classify gender (\texttt{Men} or \texttt{Women}) on the CelebA dataset.\footnote{We do not condone the use of gender predictors~\cite{hamidi2018gender}. These models merely represent our conception of the kinds of image features that would be present in a model which has severe gendered correlations.} As our \control{} pretrained models, we create three ``less biased'' ones that have been trained to classify the attributes of \texttt{Wearing} \texttt{Hat}, \texttt{Bangs}, and \texttt{Chubby}. These have been sparsely sampled to be of different saliences (how discriminable an attribute is).\footnote{We use three \control{} models of different saliences because, as we will show later, salience is a relevant factor in the transference of bias, and we want to ensure robust comparisons to a reasonable control condition.}

\begin{figure}[t!]
    \centering
    \includegraphics[width=0.44\textwidth]{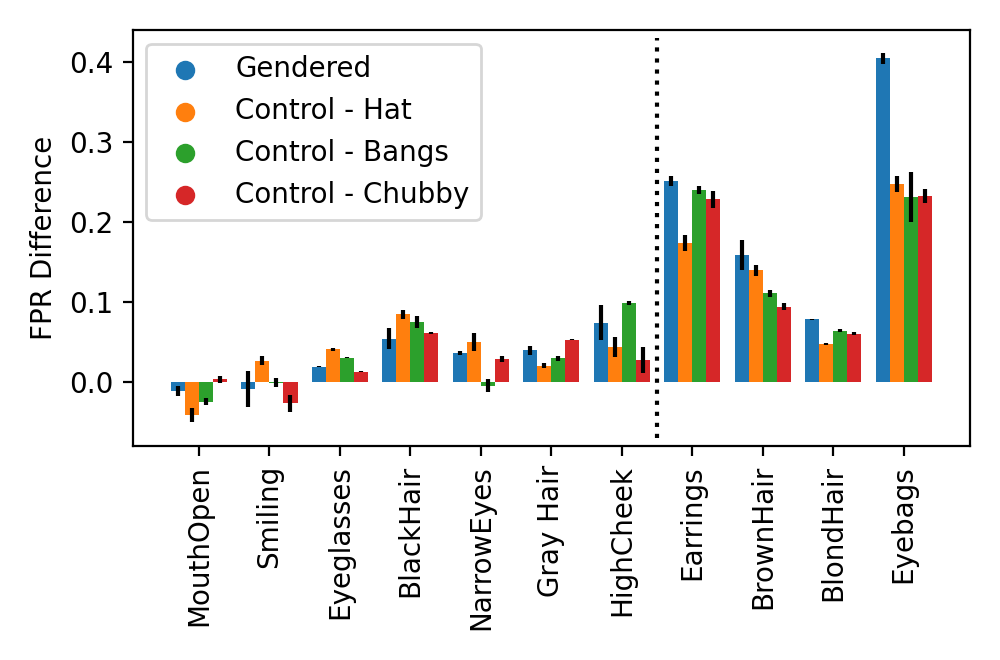}
    \caption{Across 11 CelebA attributes, the FPR difference between \texttt{Women} and \texttt{Men} of each model finetuned on a different base. The attributes are sorted by the ratio of the FPR difference between the \gendered{} model and highest \control{} model. The dotted line indicates the cutoff between the seven attributes that do not have a higher FPR difference from biased pretraining and the four attributes that do.
    }
    \label{fig:all_attributes}
\end{figure}

To understand whether finetuning on the \gendered{} model will cause more downstream bias than on a \control{} model, we compare the ratio of the FPR difference between the finetuned \gendered{} model and the most biased of the three finetuned \control{} models. As transfer learning is often done when there is insufficient data in the downstream domain, we do this on our 11 attributes using 1024 finetuning samples.
In Fig.~\ref{fig:all_attributes} we see that for 7 of our 11 attributes, finetuning on the \gendered{} model is no more biased than finetuning on any of the \control{} models; however for 4 of the attributes (\texttt{Earrings}, \texttt{Brown} \texttt{Hair}, \texttt{Blond} \texttt{Hair}, \texttt{Eyebags}), the pretrained model impacts downstream bias.

To understand what it is that differentiates the scenarios where pretrained bias does matter (4 of the 11 attributes) and where it does not (7 of the 11 attributes), we consider the relevance of three factors: (1) correlation level, (2) salience, and (3) number of finetuning samples. We show CelebA results, with COCO results in the Supplementary.



\smallsec{(1) Correlation level and (2) Salience}
Correlation level is the strength of the correlation between gender and the target attribute, and captures how useful it is for the upstream task's spurious correlation to be retained for the downstream task; salience is like the notion of discriminability, and captures whether an attribute is easier to learn in the downstream task than the spurious gender signal from the upstream task.\footnote{The notion of salience is related to simplicity bias~\cite{shah2020simplicity, valleperez2019simple, arpit2017memorization, nam2020failure} in that neural networks may be more inclined to learn certain predictive patterns due to their being more salient or simple than others.} For quantifying these two concepts we adopt the methods from prior work~\cite{zhao2017menshop, ramaswamy2021latent}. We calculate correlation level by $\frac{N(\textrm{\texttt{Women}}, \textrm{ attribute})}{N(\textrm{\texttt{Women}}, \textrm{ attribute})+N(\textrm{\texttt{Men}}, \textrm{ attribute})}$~\cite{ramaswamy2021latent} where $N(\textrm{\texttt{Women}}, \textrm{attribute})$ indicates the number of images which are labeled as \texttt{Women} and have the attribute. 
Salience is measured relative to gender in this dataset, and positive values indicate gender is more salient than the downstream attribute (e.g., \texttt{Mouth Slightly Open} is not that salient), whereas negative values indicate it is less so (e.g., \texttt{Eyeglasses} is very salient). This value is calculated by constructing a setting where gender and the attribute are exactly correlated, and the salience score is derived from which of the gender or the attribute is learned better, details in Supplementary.

For any given attribute of a particular salience, we can artificially manipulate its correlation level by selecting different subsets of the data. We do so for all attributes to be 80\%, and perform a mixed-effect regression analysis on our 22 observations (all 11 attributes of naturally different saliences at both their natural correlation level and at 80\%), with parameter estimations adjusted by the group random effects for each attribute.\footnote{We normalize correlation level to be between [0, 1], and calculate the FPR difference sign depending on which group the correlation is with.}

We find the effect size for correlation level to be 2.19 (95\% CI [.53, 3.85]) and salience to be 1.30 (95\% CI [.27, 2.34]), both with $p < .05$. 
The positive coefficient for correlation level indicates that the stronger the correlation between the target task and gender (i.e., the more it benefits a model to rely on a gendered correlation), the more it will be affected by a biased pretrained model. The positive coefficient for salience indicates that attributes of lower salience are more likely to be affected by a biased pretrained model, likely because the gender visual cue is easier to spuriously rely on compared to the true downstream task cue. 

Having seen that both correlation level and salience are able to explain part of why bias is transferred, we next investigate the effect of the number of finetuning examples.

\smallsec{(3) Finetuning number} We analyze the area under the ROC curve (AUC) and FPR difference of three pretrained models (\scratch{}, \gendered{}, \control{}) when finetuned on increasing amounts of data: $[2^4, 2^5, 2^6, ..., 2^{12}, 2^{13}, \textrm{full}=\textrm{130k}]$.
To pick the comparison \control{} from our three, we select the one with the highest AUC at $2^{10}$ for each attribute.

\begin{figure}[h]
    \centering
    \includegraphics[width=0.49\textwidth]{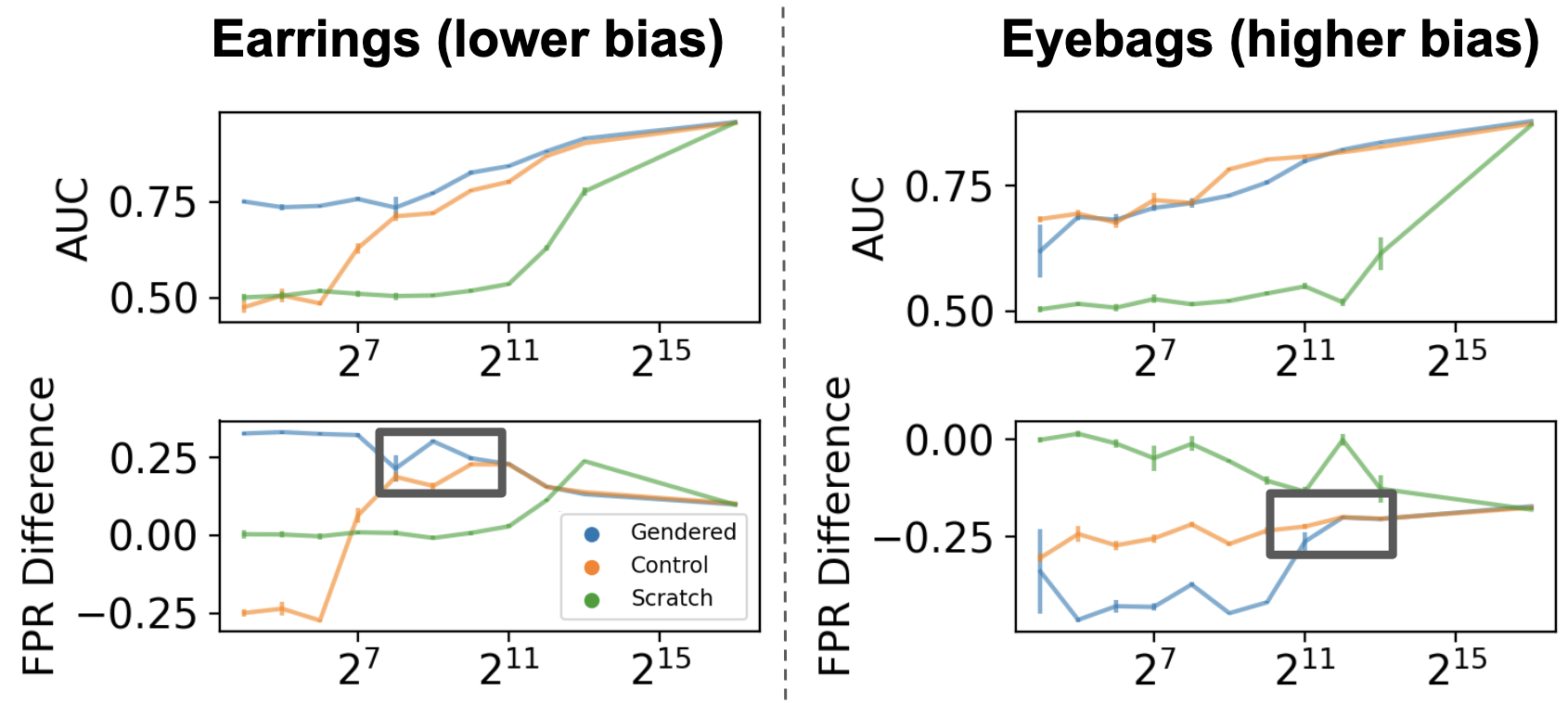}
    \caption{On two attributes from CelebA, we track the AUC and FPR difference between \texttt{Women} and \texttt{Men} for three pretrained bases (\gendered{}, \control{}, and \scratch{}) on increasing numbers of finetuning samples. 
    The \gendered{} model typically has a higher magnitude FPR difference than the \control{} model, until the two converge with sufficient finetuning samples (gray box). \texttt{Earrings}, which we saw to be less susceptible to bias from a pretrained model in Fig.~\ref{fig:all_attributes} and is of higher salience and less correlation level, achieves this at around 512 images, while \texttt{Eyebags} takes until around 4096.}
    \label{fig:nums_comp}
\end{figure}


In Fig.~\ref{fig:nums_comp} we present results on two attributes which were more biased when finetuned on the \gendered{} model compared to the \control{}: \texttt{Earrings} which had the least bias and \texttt{Eyebags} which had the most. 
Up until the full finetuning dataset is used, the \gendered{} and \control{} models which have received pretraining are able to achieve better performance than the \scratch{} model, which is in turn fairer. While with sufficient finetuning samples all three models converge to the same levels of performance and bias, we may not always have a large number of finetuning samples. \texttt{Earrings}, which has higher salience and less correlation level than \texttt{Eyebags}, requires a smaller number of finetuning samples for it not to matter in terms of FPR difference whether the \gendered{} or \control{} pretrained model was used (around 512 compared to around 4096). When the fairnesses converge, it becomes negligible which pretrained model was used. 
This is a powerful result, as it indicates we can retain potential performance gains of a pretrained model, without needing to inherit its potential biases. While using a \scratch{} model would have alleviated concerns about the bias a pretrained model, it suffers too drastically in performance in the low data regime to always be feasible. In the rest of this work, we study how we might be able to retain the performance gains of different pretrained models while not inheriting their harmful biases.



\subsection{Bias from spurious correlations can be corrected for in finetuning}
\label{sec:sc_correct}

We have now established that high correlation level, low salience, and low finetuning numbers are all relevant factors in terms of whether the bias of a pretrained model will propagate into a finetuned model.
With the exception of salience, which is hard to manually manipulate so we merely observe as a covariate, these other two factors become the levers we intervene on in our experiments going forward. 

The scenario we consider is as follows: we have a dataset for a downstream task that is small such that we would like to leverage transfer learning from a pretrained model. However, we believe the pretrained model is likely to contain biases in the form of spurious correlations that could impact the downstream task. We want to understand whether there is a way we can benefit from the performance gains that the pretrained model brings, without inheriting all of the biases.

We investigate this on two datasets, CelebA and COCO, and use five pretrained models: \scratch{}, \textbf{TorchVision}~\cite{pytorch}, \textbf{MoCo}~\cite{he2020moco}, \textbf{SimCLR}~\cite{chen2020simclr}, and \textbf{Places}~\cite{zhou2017places}. Each is finetuned on the downstream task, which has been artificially manipulated to have a correlation level of either 20\% (towards men) or 80\% (towards women), and the dataset is balanced to have equal numbers of images for the two genders. Without changing the distribution of this downstream test set, we artificially manipulate the downstream training set, i.e., finetuning dataset, across 11 increments of correlation level ([0\%, 10\%, ..., 100\%]). We then assess whether there is a manipulated setting of the finetuning dataset such that a pretrained model is able to retain most of the performance gains it brings, while having less of a FPR difference than if trained directly on the natural finetuning distribution.

\begin{figure}[t!]
    \centering
    \includegraphics[width=0.48\textwidth]{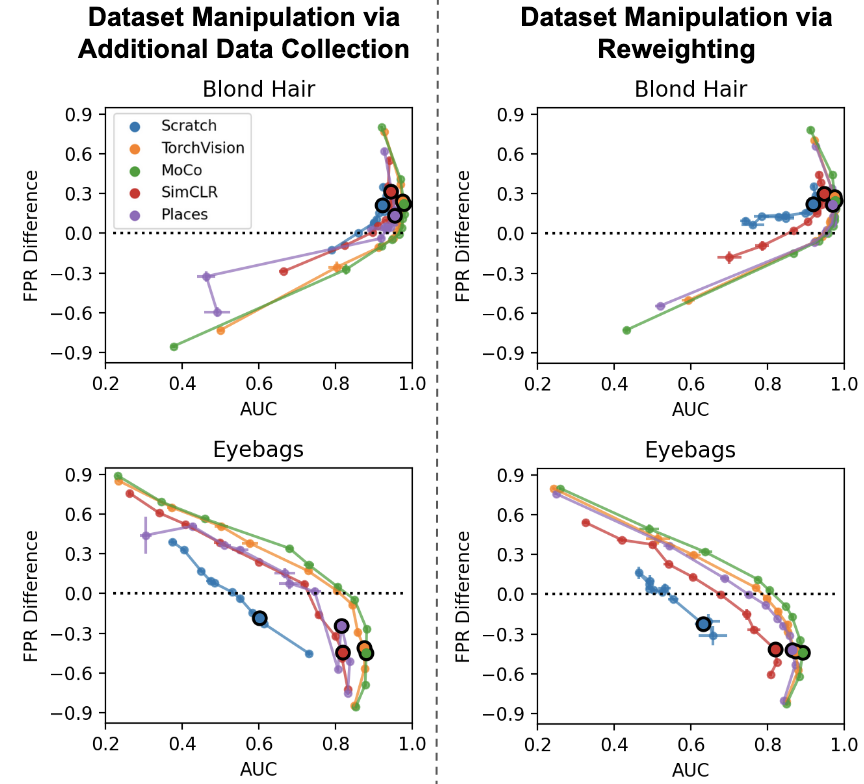}
    \caption{The performance and bias with 95\% confidence intervals of pretrained models finetuned on different versions of two downstream tasks on CelebA: \texttt{Blond Hair} (correlated with women) and \texttt{Eyebags} (correlated with men). The bolded point indicates when the finetuning distribution matches the test distribution, and all other points indicate variations on the finetuning dataset. The left column represents when additional data is collected to manipulate the correlation level while maintaining 1024 finetuning samples, and the right column represents when finetuning data is reweighted such that the correlation level changes. On both the left and right, there are versions of the finetuning dataset that allow us to retain most performance gains while improving fairness.}
    \label{fig:celeba_correlation}
\end{figure}

\smallsec{CelebA}
We use two of the four attributes most susceptible to gender biases from the previous section (\texttt{Blond} \texttt{Hair} set to correlation level 80\% and \texttt{Eyebags} set to correlation level 20\%), and show the remaining two in the Supplementary. We visualize the tradeoff between performance (AUC) and bias (FPR difference) on the left of Fig.~\ref{fig:celeba_correlation}, and find both attributes have a configuration of the finetuning dataset which is different than the test dataset such that the performance is comparable to the peak performance, but the bias is significantly lower. For example, on \texttt{Eyebags}, \textbf{MoCo} has the highest performance but also the highest bias. However, by manipulating the finetuning correlation level from 20\% to 30\% the AUC remains at $0.88$ while the FPR difference improves from $-0.46$ to $-0.27$. If we further manipulate the correlation level to 40\%, the AUC drops slightly to $0.85$, but the FPR difference improves to $-0.05$.


If we had restricted ourselves only to those models which are finetuned on a dataset that has the same correlation level as the test set (i.e., the bolded points), we would have had to compromise on performance in order to achieve lower bias. However, by manipulating the finetuning dataset such that it contains less spurious correlations than the test dataset, we can actually train a model that retains most of the performance benefits afforded to the more biased model, but also has the fairness of a less accurate model. This change is often relatively small as well, with just a 20\% shift in the correlation. In the Supplementary we show results on 128 finetuning samples, where stronger manipulations are required to decrease the FPR difference.

Of course, these gains are not free: the added cost of this approach is the need to collect additional training samples to create the new correlation level.
However, even in a constrained setting where you are unable to collect additional finetuning samples, reweighting the existing data to create an artificially less correlated dataset can bring about much of the same benefits. The right side of Fig.~\ref{fig:celeba_correlation} represents this scenario where we are unable to collect additional samples, and simply create different correlation levels by reweighting samples. 
For \texttt{Eyebags}, finetuning \textbf{MoCo} by reweighting samples such that the correlation level goes from 20\% to 30\% barely drops the performance from $0.892$ to $0.886$, while the FPR difference improves from $-0.45$ to $-0.34$. If we make the larger change to 40\% the performance drops to $0.86$, but the FPR difference is improved to $-0.17$.

\smallsec{COCO}
We investigate the two objects with the highest representation across both genders, because in this dataset the labels are more sparse and we want sufficient positive examples to be able to control the correlation level of an object with each gender. These two objects are \texttt{dining} \texttt{table}, which like the previous section we set to correlation level of 80\%, and \texttt{chair}, which we set to 20\%. We show results on additional objects in the Supplementary.

In Fig.~\ref{fig:coco_correlation} we show results on experiments with 5000 finetuning samples for the same setup as on CelebA.\footnote{We show results with 1000 finetuning examples in the Supplementary.} Because of COCO's far noisier gender cues and additional complexity due to a greater diversity of image appearances, compared to CelebA's more uniform frontal facing individuals, the trends are not as clear where each change in correlation level directly corresponds to a reduction in FPR difference. However, what remains clear is there always exists a version of the finetuning dataset which retains the high AUC of a biased pretrained model, but reduces the bias on the downstream task. For example, on \texttt{chair} the finetuned \textbf{MoCo} has an AUC of $0.81$ and FPR difference of $-0.015$; meanwhile, the finetuned \textbf{Places} has a better AUC of $0.84$ but a worse FPR difference of $-0.024$. However, instead of having to choose between lower bias or higher performance, we find that by finetuning the better-performing \textbf{Places} model on a dataset of correlation level 40\% rather than 20\%, the performance stays constant at $0.84$, while the FPR difference improves to $-0.002$! 
This indicates that careful curation of a finetuning dataset can overcome many FPR difference concerns, without compromising on the high performance that a pretrained model may be selected for. 

\begin{figure}[t!]
    \centering
    \includegraphics[width=0.48\textwidth]{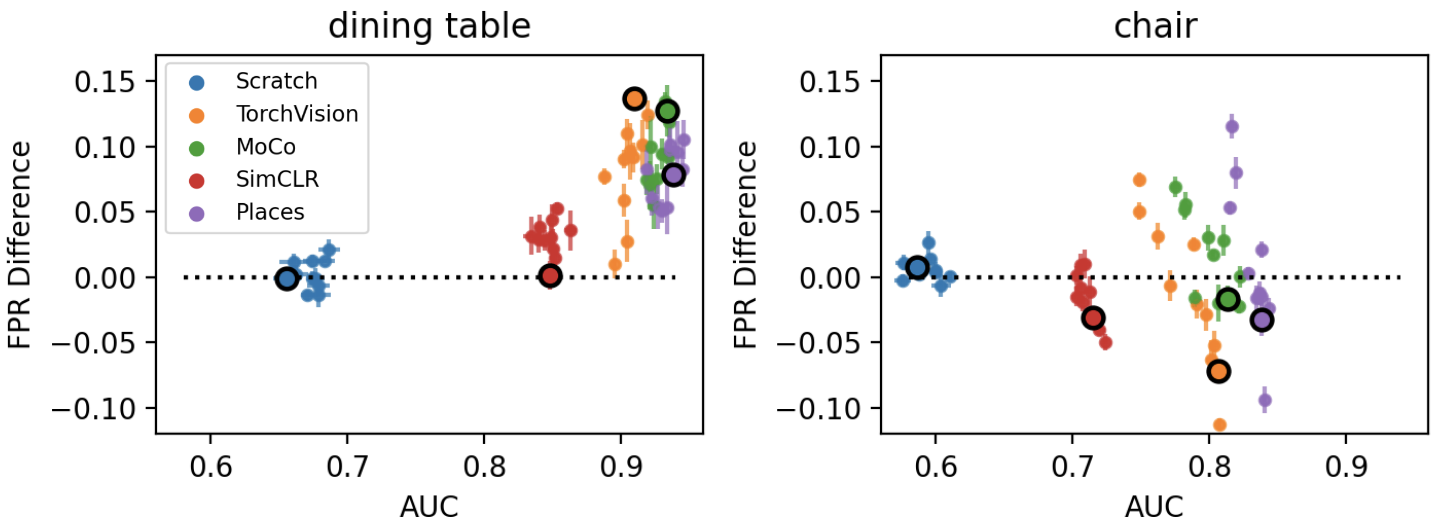}
    \caption{The performance and bias with 95\% confidence intervals of pretrained models finetuned on different versions of two downstream tasks on COCO: \texttt{dining table} (correlated with women) and \texttt{chair} (correlated with men). The bolded point indicates when the finetuning distribution matches the test distribution, and all other points indicate variations on the finetuning dataset; all models are trained on 5000 finetuning samples. In each case, there is a version of the finetuning dataset that is different in distribution from the test set which allows us to train a model which retains the high AUC of a more biased model, but has lower FPR difference.}
    \label{fig:coco_correlation}
\end{figure}

In this section we have showed how bias in the form of spurious correlations can propagate from a pretrained model to a finetuned one. However, we have also shown that manipulations to the finetuning dataset, even if they cause the finetuning dataset to deviate in distribution from the downstream test set, can actually correct for much of this bias while retaining high performance. While we have focused on the sensitive attribute of gender due to the availability of annotations, our findings are not necessarily restricted to this domain. 
Ultimately the spurious correlations we study in this section are simplified from the multi-attribute, multi-task ones that are likely to arise in the real world~\cite{wang2019balanced, li2023whac, zhao2023laundry}. We did this to make the problem tractable to study with sufficient depth, and leave further exploration for future work.
In the next section, we show similar results when bias is conceptualized as underrepresentation.



\section{Bias as Underrepresentation}
\label{sec:underrep}
Now, we consider the implications of biased pretraining when ``bias'' means that one appearance of an object is underrepresented. This is inspired from both Buolamwini and Gebru~\cite{buolamwini2018gendershades}, who showed that an underrepresentation of darker-skinned individuals led to worse classification performance for this group, and DeVries et al. \cite{devries19everyone}, who showed that appearance differences, i.e., subcategories, within an object class led to objects from countries with lower household incomes to be misclassified more often. For example, bar soap is less recognizable as soap than pump soap is. In this section, we want to understand whether a pretrained model that has only learned about one subcategory of an object (e.g., pump soap), will perform worse on the other subcategory of an object (e.g., bar soap), compared to if it had seen that subcategory during pretraining. When we find that it does, we provide insight about the level of intervention on the finetuning dataset we can perform to overcome it.

\subsection{Finetuned models do worse on subcategories underrepresented in pretrained models}
\label{sec:ur_inherit}
We consider a downstream task \texttt{Target} to be composed of two possible subcategories: \texttt{T1}, e.g., pump soap, and \texttt{T2}, e.g., bar soap. We create \textbf{Pretrain-T1} that has only been trained to classify \texttt{T1} on FairFace and \textbf{Pretrain-T2} that has only been trained to classify \texttt{T2} on FairFace. Taking \texttt{T2} to be the underrepresented subcategory that we are particularly interested in the performance of, our measure of bias is the AUC difference on \texttt{T2} between \textbf{Pretrain-T2} and \textbf{Pretrain-T1}, i.e., the performance lost by having used a pretrained model which has not seen this subcategory before.\footnote{AUC difference is between models rather than groups because different subcategories of an object may be inherently harder to classify or not, and we want to capture the relevant aspect of performance. Because FPR difference can be manipulated through post-hoc threshold changes~\cite{hardt2016equalopp}, that comparison remains between groups.
}

For our downstream task, we use the CelebA dataset, and simulate the different appearances of an object through two different attributes as the subcategories of the classification label. 
For example, ``\texttt{Light} \texttt{Hair}'' could be composed of \texttt{Blond} \texttt{Hair} and \texttt{Brown} \texttt{Hair}.
On 12 such subcategory pairings, selected to be roughly representative in terms of relative salience, we find that when we finetune on 128 images where half the positive labels are \texttt{T1} and half are \texttt{T2}, our bias measure of \texttt{T2} AUC difference is $.124\pm.023$ between the two different pretrained models. The statistically significant positive difference indicates that a finetuned \textbf{Pretrain-T1} is not able to reach the performance on \texttt{T2} that a finetuned \textbf{Pretrain-T2} is. 
Experiment details and results on the analogous three factors of correlation level, salience, and finetuning number are in the Supplementary.

\subsection{Bias from underrepresentation can be corrected for in finetuning}

We have just established that a pretrained model (e.g., \textbf{Pretrain-T1}) which has not been trained on a particular label appearance (e.g., \texttt{T2}) will perform worse on it at finetuning time than if it had been trained on it (e.g., \textbf{Pretrain-T2}). In this section, we treat the proportion of finetuning dataset that is \texttt{T1} or \texttt{T2} as our analog to correlation level, and investigate the effect of manipulating it.
It is obvious that increasing the proportion of \texttt{T2} will likely increase the performance on \texttt{T2}; however, it is not obvious what amount of proportion manipulation is required to compensate for a pretraining model not having seen a particular attribute. 

Our experimental setup is as follows: the positive labels in our downstream task are 90\% \texttt{T1} and 10\% \texttt{T2}. As we know from the previous section, if we finetune \textbf{Pretrain-T1} on this, we would not achieve as high of a \texttt{T2} AUC as if we had finetuned \textbf{Pretrain-T2}. However, it is unlikely we have access to \textbf{Pretrain-T2} because existing pretrained models are often biased towards having been trained on, for example, geographically homogenous images~\cite{devries19everyone, shankar2017geodiversity, revisetool_extended}. Thus, in our setting in which we are actually able to compare to \textbf{Pretrain-T2}, we ask what proportion of the finetuning dataset needs to be \texttt{T2} such that the \textbf{Pretrain-T1} we actually have can achieve comparable performance. 
In other words, the amount of change to the finetuning dataset that correlates with having used an entirely different pretrained model to begin with.

\smallsec{CelebA}
On our 12 attribute pairs we find that for 128 finetuning samples, manipulating the proportion of \texttt{T2} from just 10\% to 30\% brings the performance of \textbf{Pretrain-T1} to that of \textbf{Pretrain-T2} in 6 of our 12 cases, and that manipulation on for 1024 finetuning samples does so in 8 of our 12 cases.



\smallsec{Dollar Street and COCO} 
Applying this insight from CelebA that relatively minor manipulations to the proportion of the dataset from underrepresented subcategories can significantly impact the performance of those subcategories, we now turn to the more complex and realistic object datasets of Dollar Street and COCO.

We consider the task of recognizing 15 objects in Dollar Street that have corresponding objects in COCO. Although the object classes are the same between datasets, their visual distribution is different as COCO images largely come from only higher-income regions~\cite{devries19everyone} whereas Dollar Street was collected to be more geographically diverse. Nevertheless COCO images are more plentiful; to simulate this, we consider a finetuning dataset of 128 images where 90\% are from COCO and 10\% from Dollar Street. 

We use two pretrained models, named after the dataset each is trained on: \textbf{ImageNet}~\cite{imagenet}, where the training data is more similar to the COCO distribution, and \textbf{GeoDE}~\cite{ramaswamy2023geode}, which is trained on a newer and more geographically diverse dataset. As expected, finetuning \textbf{GeoDE} achieves a higher accuracy (where a prediction is correct if the object is one of the top-5 predicted labels~\cite{devries19everyone}) of 13.1\% on Dollar Street compared to finetuning \textbf{ImageNet} with 8.4\%. 

\begin{figure}[t!]
    \centering
    \includegraphics[width=0.35\textwidth]{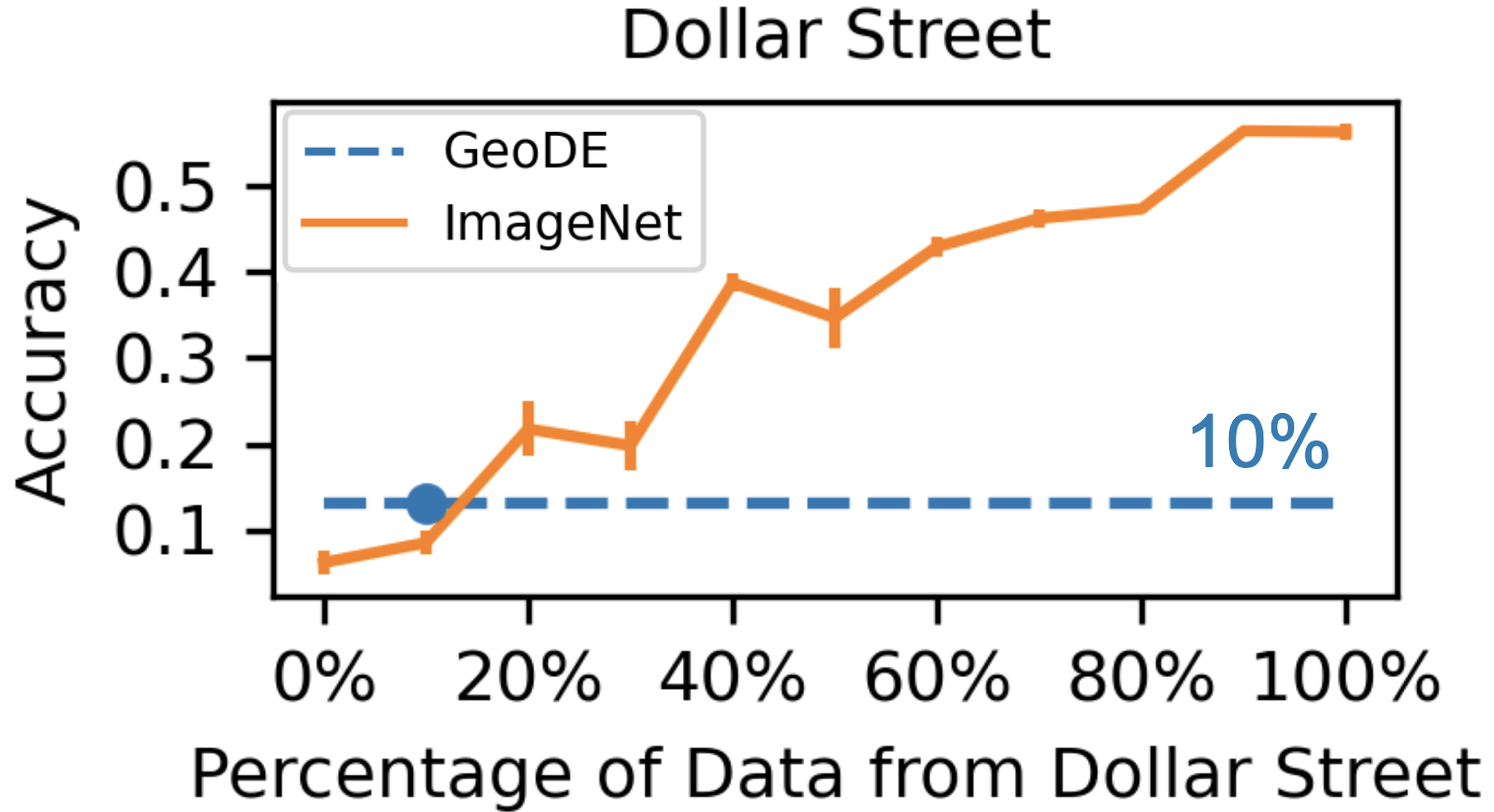}
    \caption{The performance of an \textbf{ImageNet}-pretrained model~\cite{imagenet} at recognizing 15 objects in the geographically diverse Dollar Street~\cite{rojas2022dollarstreet} dataset, as the proportion of finetuning images from Dollar Street itself (more geographically diverse) and COCO~\cite{li2014coco} (less so) is manipulated. Only a minor change in finetuning proportion is required for the \textbf{ImageNet}-pretrained model to match the performance of a \textbf{GeoDE}-pretrained model, which was trained on a dataset curated to be more geographically diverse~\cite{ramaswamy2023geode}.  
    }
    \label{fig:dollarprops}
\end{figure}

However, what we ultimately want to investigate is how much investing in a better finetuning dataset can help overcome the problem (Fig.~\ref{fig:dollarprops}). Thus, we manipulate the finetuning dataset (simulating the collection of more Dollar Street-like images, while keeping the overall finetuning number the same), and observe that with just 20\% rather than 10\% of images coming from Dollar Street, \textbf{ImageNet} is able to outperform the performance of the \textbf{GeoDE} baseline with an accuracy of 21.7\%.

\section{Discussion}

Across both operationalizations of bias, as \textit{spurious correlations} and \textit{underrepresentation}, we find that bias from a pretrained model can propagate into a finetuned one. When bias is in the form of a \textit{spurious correlation}, this is especially likely when the downstream task has a high correlation level with the sensitive attribute, is of low salience, and there is a small number of finetuning samples. For those creating pretrained models, responsibility needs to be taken to clearly document the data the model was trained on, as well as known biases it may have~\cite{gebru2021datasheets, mitchell2019modelcards}.

However, across both of these conceptualizations of bias, we also find that the distribution of the finetuning dataset has the ability to counteract much of the bias a pretrained model may bring. When the bias we are concerned with is a \textit{spurious correlation}, this entails manipulating the correlation level in the finetuning dataset, and when the bias is \textit{underrepresentation}, this entails manipulating the proportion of positive labels which are of the underrepresented group. Necessarily, all of these manipulations will require additional data collection efforts. However, the efforts are significantly lower than would be required to intervene on the massive pretraining dataset. 
It is notable that these manipulations allow us to retain most of the performance gains that a biased pretrained base might bring. Acting on these findings will require careful care and consideration in the collection of finetuning datasets, at times even over-correcting for a particular bias that may be present in the downstream test set. 
Especially given that most settings we investigate are those with smaller finetuning datasets, these being the scenarios where pretrained models are most necessary, we should expect and be willing to put in thoughtful work in curating finetuning datasets~\cite{scheuerman2021datasets, paullada2021discontents, sambasivan2021datacascades}.
Data collection can often be extractive and violate privacy \cite{jo2020archives, prabhu21pyrrhic}, but there are ways in which it can be done more consensually~\cite{ramaswamy2023geode}. Given that the relevant harms are usually context-specific and hard to conceptualize upstream of the downstream task, this lends further support to the notion that the finetuning dataset makes for both a practical and efficient point of intervention in bias mitigation.

\subsection{Limitations}
There are a number of limitations that qualify the generalizability of our findings. All experiments are conducted on a ResNet50, which uses a convolutional neural network architecture. Prior work validates that a subset of their experiments which are similar to ours have comparable results across model architectures, and we build on this intuition~\cite{salman2022transfer}. 
However, as the size of pretrained models increases dramatically, we leave for future work further exploration.
Additionally, we finetune all of the model's weights rather than freezing some.

Most significantly, we only conceive of two possible operationalizations of bias in this work: spurious correlations and underrepresentation. Other types of biases (e.g., stereotypical representations) will likely lead to different transferrence properties.
In our technically defined notions of bias, we have also left out of scope considerations such as NSFW content, privacy issues, and other harms of pretraining datasets~\cite{birhane2021multimodal, prabhu21pyrrhic}. Irrespective of their downstream effects, pretraining on these images are inherently harmful because of impacts such as on the individuals potentially tasked with annotating the data. Ultimately in this work, we explored very directed interventions on the finetuning dataset to target specific forms of bias. However, likely any downstream task will have many different kinds of bias that are relevant, and potentially even at odds~\cite{wang2022representational}. We leave for future work how to balance potentially conflicting tensions in order to curate more suitable finetuning datasets.

\section{Conclusion}
Finetuning on top of pretrained models is a powerful way to train models on domains where we have less data. In this work, we conceive of bias as \textit{spurious correlations} or \textit{underrepresentation} and show that biases of either form can transfer from pretrained models to finetuned ones. However, we also affirmatively show that targeted manipulations of the finetuning dataset can counteract this bias transferrance, allowing us to retain performance gains that certain pretrained models with bias may bring, without compromising on fairness concerns. These dual findings indicate that while interventions on the pretrained model to ensure less bias in the features are certainly useful, more effective interventions can be performed by manipulating the finetuning dataset itself. The benefit of manipulations at this juncture is there will also likely be a better understanding of the application-specific harms for which dataset intervention can be targeted. This will require a careful, participatory, and deliberative curation of the finetuning dataset.

\section*{Acknowledgements}
This material is based upon work supported by the National Science Foundation under Grant No. 1763642, Grant No. 2112562, Grant No. 2145198, and Graduate Research Fellowship to AW. Any opinions, findings, and conclusions or recommendations expressed in this material are those of the author(s) and do not necessarily reflect the views of the National Science Foundation. We thank Allison Chen, Vikram V. Ramaswamy, and Shruthi Santhanam for feedback.

\clearpage

{\small
\bibliographystyle{ieee_fullname}
\bibliography{references}
}

\clearpage

\appendix


\section{Salience Calculation}
Following from Ramaswamy et al.~\cite{ramaswamy2021latent}, we calculate the relative salience between two attributes, \texttt{A} and \texttt{B} in the following manner. We sample a dataset from CelebA where half the attributes have both \texttt{A} and \texttt{B}, and half have neither. We then train a model to perform binary classification on this dataset. Then, we test this model on a test set composed of images equally sampled from the four possible conjunctions of the two attributes. We calculate the AUC on attribute \texttt{A} and \texttt{B} independently, and take the difference between their performance to indicate which attribute is more salient. We repeat this experiment on a second training dataset skewed for the inverse of \texttt{A}, i.e., the two halves are \texttt{A} and not-\texttt{B}, and not-\texttt{A} and \texttt{B}.

\section{Correlation Level, Salience, Number of Finetuning Samples on COCO}
In Sec.~\ref{sec:sc_inherit} we demonstrated how on the CelebA dataset the three factors of correlation level, salience, and number of finetuning images impact whether bias from a pretrained model in the form of a \textit{spurious correlation} will propagate into a finetuned one. Here, we show how these same factors are relevant on the more complex COCO dataset.

In creating our pretrained models, we use the OpenImages dataset~\cite{kuznetsova2020openimages}, which has 600 labels such as \texttt{Ladder} and \texttt{Carrot} that are annotated by a combination of humans and machines. For gender we follow from prior work~\cite{zhao2017menshop, revisetool_extended} and derive these labels based on the presence of gendered labels in the original dataset.\footnote{Schumann et al.~\cite{schumann2021miap} has collected a set of more inclusive gender presentaiton labels on this dataset, but we did not use them because the subset these labels exist for was not large enough for our purposes of pretraining.} We use the subset of this 
dataset which contains people of only one annotated gender. 

Like in CelebA, we use two types of pretrained models: \gendered{} and \control{}. In this setting, our \gendered{} model is pretrained to classify binary gender on OpenImages. Our \control{} model is trained to classify \texttt{ outdoor parks} or not on OpenImages. These scene labels come from a Places scene classifier~\cite{zhou2017places}.

We work with four versions of COCO as our downstream task. Two have varying levels of correlation strength, and two have varying levels of salience. We would expect that on the dataset with more correlation strength compared to less, there would be a bias difference between the two pretrained models; we would not expect as large of a difference on the dataset with less correlation strength. The same hypothesis holds for the two datasets of different salience. To create two versions of COCO that have different levels of correlation strength between the target task (i.e., objects) and sensitive attribute (i.e., gender) we train a logistic regression model to predict gender from a binary vector representing all of the objects present in an image. We then sort all images in the dataset by those most correctly classified by this model to those least correctly classified. We split the images in half to create two datasets, the first we call ``Skew\_MoreBias'' (object presence is highly correlated with gender) and the second set ``Skew\_LessBiass'' (object presence is less correlated with gender). To create two versions of COCO where the salience of the target task and sensitive attribute differ, we blur all of the objects to create ``Salience\_MoreBias'' and blur all of the people to create ``Salience\_LessBias.'' As our measure of bias we use the directional bias amplification measure ($\textrm{BiasAmp}_{\rightarrow}$) from Wang and Russakovsky~\cite{wang2021biasamp}.

\begin{figure}[h]
    \centering
    \includegraphics[width=0.49\textwidth]{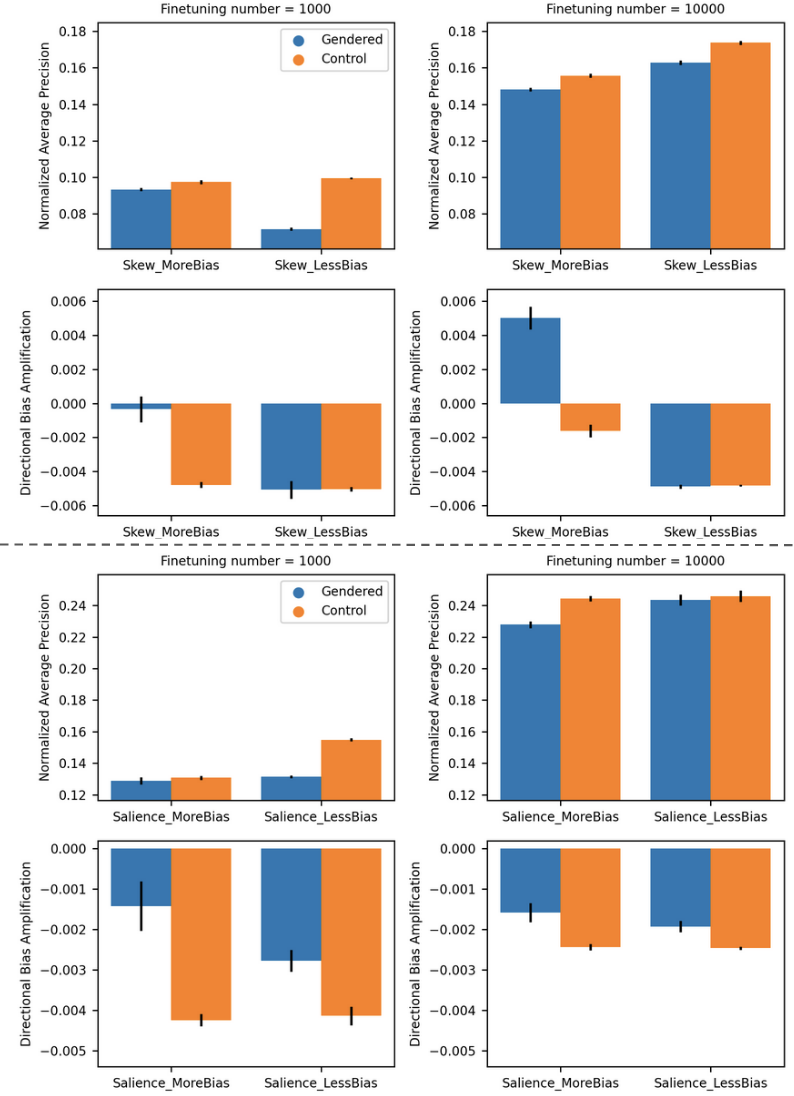}
    \caption{We present results for the performance (normalized average precision) and bias (directional bias amplification) of finetuned models pretrained on two bases: \textbf{Gendered} and \textbf{Control}. We show results on four different datasets, ``Skew\_MoreBias'' compared to ``Skew\_LessBiass,'' and ``Salience\_MoreBias'' compared to ``Salience\_LessBias.''}
    \label{fig:coco_salskew}
\end{figure}

In Fig.~\ref{fig:coco_salskew} we show results from finetuning our \gendered{} and \control{}  pretrained models on all four variations of the COCO dataset on both 1,000 and 10,000 finetuning samples. When we first compare the results of the two pretrained models on ``Skew\_MoreBias'' and ``Skew\_LessBias'' we see that on the former dataset it makes a difference which pretrained model is used, while for the latter it does not. Somewhat unexpectedly, when we increas the number of finetuning samples from 1,000 to 10,000, the difference between the two pretraining bases increases rather than decreases, as we saw in CelebA. We hypothesize this is because the dataset we have created is so skewed that it benefits the model significantly to continue to learn the spurious correlations, even as the finetuning number has increased. 

We see the same results in Fig.~\ref{fig:coco_salskew} for salience where on ``Salience\_MoreBias'' there is a higher difference in directional bias amplification between the two different pretrained bases. However, here we see this gap reduces with additional finetuning samples.

\section{Additional Results from ``4.2 Bias from spurious correlations can be corrected for in finetuning''}

\begin{figure}[h]
    \centering
    \includegraphics[width=0.49\textwidth]{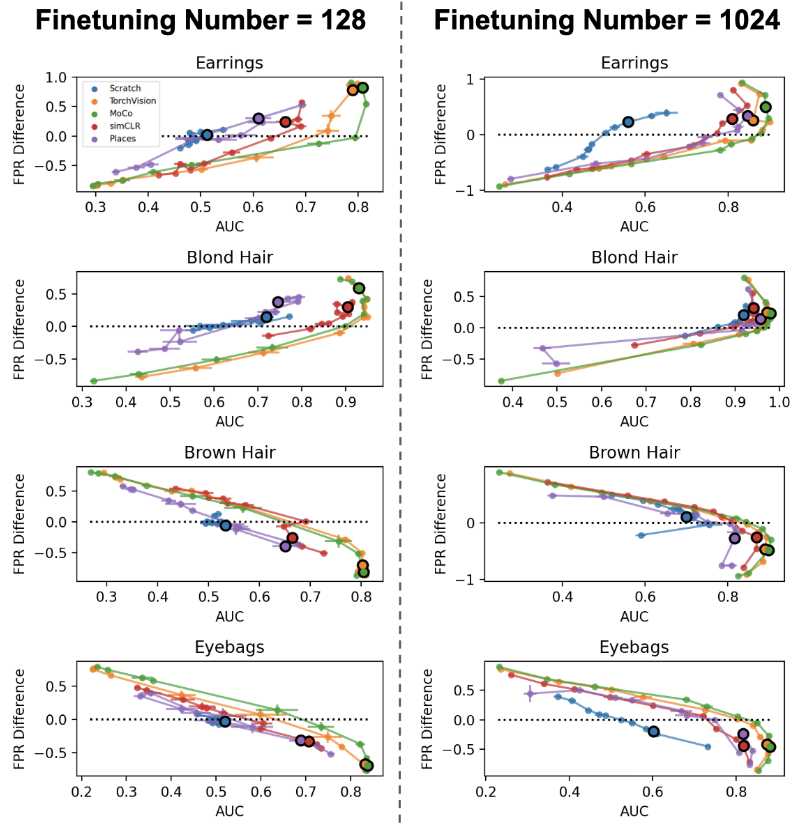}
    \caption{The performance and fairness with 95\% confidence intervals of pretrained models finetuned on different versions of four downstream tasks on CelebA: \texttt{Earrings} and \texttt{Blond} \texttt{Hair} (correlated with women) and \texttt{Brown} \texttt{Hair} and \texttt{Eyebags} (correlated with men). The bolded point indicates when the finetuning distribution matches the test distribution, and all other points indicate variations on the finetuning dataset. There are versions of the finetuning dataset that allow us to retain performance gains and improve fairness.}
    \label{fig:supp_celebacorr}
\end{figure}

In Sec.~\ref{sec:sc_correct} we showed results from manipulating the correlation level of the finetuning dataset on both CelebA and COCO. Here, we show additional results on a larger set of downstream tasks for each dataset, as well as different numbers of finetuning samples. In Fig.~\ref{fig:supp_celebacorr} we show results with 128 and 1024 finetuning samples on the four CelebA attributes which exhibit bias transfer from the pretrained models to the finetuned one. Just like the results we show in the main text, there is a version of each finetuning dataset such that the distribution has a different correlation level than the test dataset, but the performance is retained while fairness improves. 

\begin{figure}[h]
    \centering
    \includegraphics[width=0.49\textwidth]{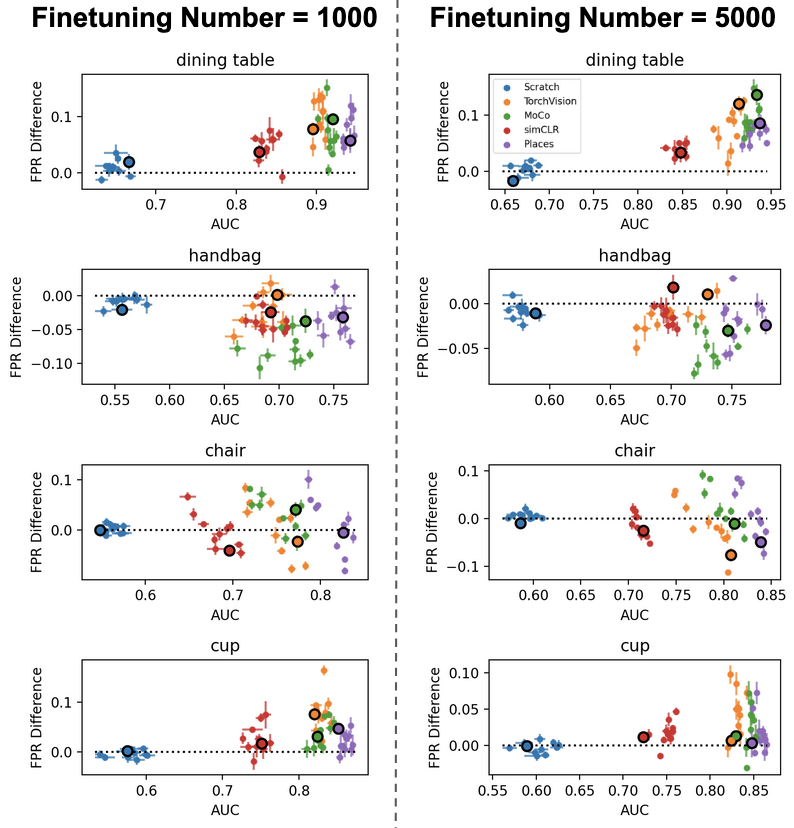}
    \caption{The performance and fairness with 95\% confidence intervals of pretrained models finetuned on different versions of four downstream tasks on COCO: \texttt{dining} \texttt{table} and \texttt{handbag} (correlated with women) and \texttt{chair} and \texttt{cup} (correlated with men). The bolded point indicates when the finetuning distribution matches the test distribution, and all other points indicate variations on the finetuning dataset. We can see that across both numbers of finetuning there are versions of the finetuning dataset that allow us to retain performance gains and improve fairness.}
    \label{fig:supp_cococorr}
\end{figure}

In Fig.~\ref{fig:supp_cococorr} we show results with 1000 and 5000 finetuning samples on the four COCO objects which are most represented with both genders. Again, like in the main text, we see that there exists versions of the finetuning dataset that allow us to preserve the high performance of a more biased model while decreasing the bias.

\section{Additional Results from ``5.1 Finetuned models do worse on subcategories underrepresented in pretrained models''}

In Sec.~\ref{sec:ur_inherit} we showed on CelebA that finetuned models inherit biases in the form of underrepresentation from pretrained models. Here, we provide further details about our experimental setup, as well as more detailed results disaggregated by correlation level, salience, and finetuning number.

As we had described in our setup, we consider our downstream task \texttt{Target} to be composed of two possible subcategories: \texttt{T1} and \texttt{T2}. We have two possible pretrained models: \textbf{Pretrain-T1} that has only been trained to classify \texttt{T1}, and \textbf{Pretrain-T2} that has only been trained to classify \texttt{T2}. Our measure of bias is AUC on \texttt{T2} between \textbf{Pretrain-T2} and \textbf{Pretrain-T1}.

For any instantiation of \texttt{T1} and \texttt{T2} using CelebA attributes, \textbf{Pretrain-T1} and \textbf{Pretrain-T2} are trained on their respective attributes on the FairFace dataset. FairFace does not contain attribute labels, so these are labeled by our best classifier which was originally trained on CelebA. While imperfect, we believe this will still provide sufficient training signal for each pretrained model.

We consider three relevant factors which are analogous to those we considered in Sec.~\ref{sec:sc_inherit} of correlation level, salience, and finetuning number. For correlation level, we consider the proportion of positive labels that are in subcategory \texttt{T1} as compared to \texttt{T2}. For salience, we consider the relative salience of \texttt{T1} and \texttt{T2}. Finetuning number remains the same.
In picking attribute pairs to use as \texttt{T1} and \texttt{T2}, we sample from three discretized types of salience relationships: \texttt{T1} is more salient than \texttt{T2}, \texttt{T1} is equally salient to \texttt{T2}, and \texttt{T1} is less salient than \texttt{T2}. We arbitrarily pick four attribute pairs from each category, and thus look at 12 pairs.

In establishing that finetuned models can inherit biases of underrepresentation from pretrained models, we consider when the downstream dataset is 50\% \texttt{T1} and 50\% \texttt{T2}; the same results on additional proportions are shown in full in Tbl.~\ref{tbl:t1vst2}. We do not find clear trends in performance difference on \texttt{T2} across the three possible salience relationships, but we do for finetuning number. Across the 12 attribute pairs when we finetune on 128 images, the difference in AUC is $.124\pm.023$, whereas when we finetune on 1024 images, the difference is $.036\pm.007$. As expected, increasing numbers of finetuning samples erodes the difference between the different pretrained bases. However, in both cases there is a statistically significant positive difference indicating that a finetuned \textbf{Pretrain-T1} is not able to reach the performance on \texttt{T2} that a finetuned \textbf{Pretrain-T2} is. Even though we do not observe any difference between the different settings of salience, we continue all experiments in the main text across these 12 pairings for generalizability.

\begin{table}[t!]
\caption{The top table represents when the finetuning number is 128, and the bottom when it is 1024.}
\label{tbl:t1vst2}
\begin{minipage}{.47\linewidth}
\begin{tabular}{|>{\centering\arraybackslash}p{1.7cm}|>{\centering\arraybackslash}p{1.7cm}|>{\centering\arraybackslash}p{1.7cm}|>{\centering\arraybackslash}p{1.7cm}|}
\hline
Correlation Strength / Salience & T1 less than T2 & T1 equal to T2 & T1 more than T2  \\ \hline
10\% & $.17 \pm .06$ &  $.06 \pm .03$ & $.13 \pm .09$\\ \hline
50\% & $.15 \pm .07$ & $.07 \pm .02$ & $.15 \pm .09$ \\ \hline
90\% & $.10 \pm .06$ & $.07 \pm .03$ & $.07 \pm .09$  \\ \hline
\end{tabular}
   \end{minipage}%
\\
\begin{minipage}{.47\linewidth}
\begin{tabular}{|>{\centering\arraybackslash}p{1.7cm}|>{\centering\arraybackslash}p{1.7cm}|>{\centering\arraybackslash}p{1.7cm}|>{\centering\arraybackslash}p{1.7cm}|}
\hline
Correlation Strength / Salience & T1 less than T2 & T1 equal to T2 & T1 more than T2 \\ \hline
10\% & $.04 \pm .02$ & $.00 \pm .00$ & $.03 \pm .02$ \\ \hline
50\% & $.03 \pm 0.02$ & $.03 \pm .01$ & $.05 \pm 0.03$ \\ \hline
90\% & $.05 \pm 0.04$ & $.02 \pm .01$ & $.14 \pm 0.05$ \\ \hline
\end{tabular}
   \end{minipage}%
   \end{table}


\end{document}